\documentclass[10pt,twocolumn,letterpaper]{article}

\usepackage{cvpr}              

\usepackage[accsupp]{axessibility}

\usepackage{multirow}
\usepackage[table,xcdraw]{xcolor}
\usepackage[table,xcdraw]{xcolor}
\usepackage{algorithmic}
\usepackage{algorithm}
\usepackage{pifont} 

\usepackage{booktabs,caption,subcaption,siunitx,xcolor,makecell}
\definecolor{cvprblue}{rgb}{0.21,0.49,0.74}
\usepackage[pagebackref,breaklinks,colorlinks,allcolors=cvprblue]{hyperref}

\def\paperID{*****} 
\def\confName{CVPR}
\def\confYear{2026}

\title{Dual Strategies for Test-Time Adaptation}

\author{
Nam Nguyen Phuong$^{1}$ \hspace{1ex}
Duc Nguyen The Minh$^{1}$ \hspace{1ex}
Phi Le Nguyen$^{1*}$ \hspace{1ex}
Ehsan Abbasnejad$^{2}$ \hspace{1ex}
Minh Hoai$^{3}$\\
\\
$^{1}$ Institute for AI Innovation and Societal Impact, Hanoi University of Science and Technology \\
$^{2}$ Monash University \quad
$^{3}$ Adelaide University \quad 
$^*$ Corresponding author
}
\begin{document}
\maketitle
\def\mA{\mathcal{A}}
\def\mB{\mathcal{B}}
\def\mC{\mathcal{C}}
\def\mD{\mathcal{D}}
\def\mE{\mathcal{E}}
\def\mF{\mathcal{F}}
\def\mG{\mathcal{G}}
\def\mH{\mathcal{H}}
\def\mI{\mathcal{I}}
\def\mJ{\mathcal{J}}
\def\mK{\mathcal{K}}
\def\mL{\mathcal{L}}
\def\mM{\mathcal{M}}
\def\mN{\mathcal{N}}
\def\mO{\mathcal{O}}
\def\mP{\mathcal{P}}
\def\mQ{\mathcal{Q}}
\def\mR{\mathcal{R}}
\def\mS{\mathcal{S}}
\def\mT{\mathcal{T}}
\def\mU{\mathcal{U}}
\def\mV{\mathcal{V}}
\def\mW{\mathcal{W}}
\def\mX{\mathcal{X}}
\def\mY{\mathcal{Y}}
\def\mZ{\mathcal{Z}} 

\def\bbN{\mathbb{N}} 
\def\bbR{\mathbb{R}} 
\def\bbP{\mathbb{P}} 
\def\bbQ{\mathbb{Q}} 
\def\bbE{\mathbb{E}}

\def\1n{\mathbf{1}_n}
\def\0{\mathbf{0}}
\def\1{\mathbf{1}}

\def\A{{\bf A}}
\def\B{{\bf B}}
\def\C{{\bf C}}
\def\D{{\bf D}}
\def\E{{\bf E}}
\def\F{{\bf F}}
\def\G{{\bf G}}
\def\H{{\bf H}}
\def\I{{\bf I}}
\def\J{{\bf J}}
\def\K{{\bf K}}
\def\L{{\bf L}}
\def\M{{\bf M}}
\def\N{{\bf N}}
\def\O{{\bf O}}
\def\P{{\bf P}}
\def\Q{{\bf Q}}
\def\R{{\bf R}}
\def\S{{\bf S}}
\def\T{{\bf T}}
\def\U{{\bf U}}
\def\V{{\bf V}}
\def\W{{\bf W}}
\def\X{{\bf X}}
\def\Y{{\bf Y}}
\def\Z{{\bf Z}}

\def\a{{\bf a}}
\def\b{{\bf b}}
\def\c{{\bf c}}
\def\d{{\bf d}}
\def\e{{\bf e}}
\def\f{{\bf f}}
\def\g{{\bf g}}
\def\h{{\bf h}}
\def\i{{\bf i}}
\def\j{{\bf j}}
\def\k{{\bf k}}
\def\l{{\bf l}}
\def\m{{\bf m}}
\def\n{{\bf n}}
\def\o{{\bf o}}
\def\p{{\bf p}}
\def\q{{\bf q}}
\def\r{{\bf r}}
\def\s{{\bf s}}
\def\t{{\bf t}}
\def\u{{\bf u}}
\def\v{{\bf v}}
\def\w{{\bf w}}
\def\x{{\bf x}}
\def\y{{\bf y}}
\def\z{{\bf z}}

\def\balpha{\mbox{\boldmath{$\alpha$}}}
\def\bbeta{\mbox{\boldmath{$\beta$}}}
\def\bdelta{\mbox{\boldmath{$\delta$}}}
\def\bgamma{\mbox{\boldmath{$\gamma$}}}
\def\blambda{\mbox{\boldmath{$\lambda$}}}
\def\bsigma{\mbox{\boldmath{$\sigma$}}}
\def\btheta{\mbox{\boldmath{$\theta$}}}
\def\bomega{\mbox{\boldmath{$\omega$}}}
\def\bxi{\mbox{\boldmath{$\xi$}}}
\def\bnu{\mbox{\boldmath{$\nu$}}}                                  
\def\bphi{\mbox{\boldmath{$\phi$}}}
\def\bmu{\mbox{\boldmath{$\mu$}}}

\def\bDelta{\mbox{\boldmath{$\Delta$}}}
\def\bOmega{\mbox{\boldmath{$\Omega$}}}
\def\bPhi{\mbox{\boldmath{$\Phi$}}}
\def\bLambda{\mbox{\boldmath{$\Lambda$}}}
\def\bSigma{\mbox{\boldmath{$\Sigma$}}}
\def\bGamma{\mbox{\boldmath{$\Gamma$}}}
                                  
\newcommand{\myprob}[1]{\mathop{\mathbb{P}}_{#1}}

\newcommand{\nam}[1]{\textcolor{orange}{#1}}
\newcommand{\myexp}[1]{\mathop{\mathbb{E}}_{#1}}

\newcommand{\mydelta}[1]{1_{#1}}

\newcommand{\myminimum}[1]{\mathop{\textrm{minimum}}_{#1}}
\newcommand{\mymaximum}[1]{\mathop{\textrm{maximum}}_{#1}}    
\newcommand{\mymin}[1]{\mathop{\textrm{minimize}}_{#1}}
\newcommand{\mymax}[1]{\mathop{\textrm{maximize}}_{#1}}
\newcommand{\mymins}[1]{\mathop{\textrm{min.}}_{#1}}
\newcommand{\mymaxs}[1]{\mathop{\textrm{max.}}_{#1}}  
\newcommand{\myargmin}[1]{\mathop{\textrm{argmin}}_{#1}} 
\newcommand{\myargmax}[1]{\mathop{\textrm{argmax}}_{#1}} 
\newcommand{\myst}{\textrm{s.t. }}

\newcommand{\denselist}{\itemsep -1pt}
\newcommand{\sparselist}{\itemsep 1pt}

\definecolor{pink}{rgb}{0.9,0.5,0.5}
\definecolor{purple}{rgb}{0.5, 0.4, 0.8}   
\definecolor{gray}{rgb}{0.3, 0.3, 0.3}
\definecolor{mygreen}{rgb}{0.2, 0.6, 0.2}

\newcommand{\cyan}[1]{\textcolor{cyan}{#1}}
\newcommand{\blue}[1]{\textcolor{blue}{#1}}
\newcommand{\magenta}[1]{\textcolor{magenta}{#1}}
\newcommand{\pink}[1]{\textcolor{pink}{#1}}
\newcommand{\green}[1]{\textcolor{green}{#1}} 
\newcommand{\mygreen}[1]{\textcolor{mygreen}{#1}}    
\newcommand{\purple}[1]{\textcolor{purple}{#1}}       

\definecolor{greena}{rgb}{0.4, 0.5, 0.1}
\newcommand{\greena}[1]{\textcolor{greena}{#1}}

\definecolor{bluea}{rgb}{0, 0.4, 0.6}
\newcommand{\bluea}[1]{\textcolor{bluea}{#1}}
\definecolor{reda}{rgb}{0.6, 0.2, 0.1}
\newcommand{\reda}[1]{\textcolor{reda}{#1}}

\def\changemargin#1#2{\list{}{\rightmargin#2\leftmargin#1}\item[]}
\let\endchangemargin=\endlist

\newcommand{\mduc}[1]{\textcolor{blue}{#1}}

\newcommand{\mhoai}[1]{{\color{magenta}\textbf{[MH: #1]}}}

\newcommand{\mtodo}[1]{{\color{red}$\blacksquare$\textbf{[TODO: #1]}}}
\newcommand{\myheading}[1]{\vspace{0.5ex}\noindent \textbf{#1}}
\newcommand{\htimesw}[2]{\mbox{$#1$$\times$$#2$}}


%
%
%

\newcommand{\Sref}[1]{Sec.~\ref{#1}}
\newcommand{\Eref}[1]{Eq.~(\ref{#1})}
\newcommand{\Fref}[1]{Fig.~\ref{#1}}
\newcommand{\Tref}[1]{Table~\ref{#1}}

\begin{abstract}

Conventional test-time adaptation (TTA) approaches typically adapt the model using only a small fraction of test samples, often those with low-entropy predictions, thereby failing to fully leverage the available information in the test distribution. This paper introduces DualTTA, a novel framework that improves performance under distribution shifts by utilizing a larger and more diverse set of test samples. DualTTA identifies two distinct groups: one where the model's predictions are likely consistent with the underlying semantics, and another where predictions are likely incorrect. For the first group, it minimizes prediction entropy to reinforce reliable decisions; for the second, it maximizes entropy to suppress overconfident errors and unlearn spurious behavior. These groups are adaptively selected using a new reliability criterion that measures prediction stability under both semantic-preserving and semantic-altering transformations, addressing the limitations of purely entropy-based selection. We further provide theoretical analysis and empirical justification showing that our approach enables a tighter separation between reliable and unreliable samples - in the context of their suitability for adaptation - leading to provably more effective model updates. 
The source code is available at 
\url{https://github.com/namk65hust/DualTTA}.

\end{abstract}    
\section{Introduction}
\label{sec:intro}
Deep learning models can struggle when training and test distributions differ, a challenge known as distribution shift. For example, in many computer vision applications, this occurs when statistical properties of the source and target data diverge due to factors like image corruption, lighting changes, or adverse weather \cite{hendrycks2019benchmark,koh2021wilds}. As deep networks are highly sensitive to such shifts, adapting models to test scenarios is crucial for maintaining reliability in real-world deployments. To address this issue, various approaches have been proposed, including domain generalization \cite{zhou2022domain, zhou2021domain,tran2025conststyle}, domain adaptation \cite{farahani2020brief, sun2015survey}, test-time training \cite{liu2021ttt, sun2020test}, and Test-Time Adaptation (TTA) \cite{chen2022contrastive, liang2024comprehensive, lee2024entropy, niu2022efficient, niu2023towards, wang2021tent, m_Zhang-Hoai-CVPR23, m_Zhang-etal-BMVC24}. This paper focuses on TTA, a practical setting in which a trained  model must adapt to an unseen target domain during inference without access to test data labels or the original training data.

\begin{figure}[t]
    \centering
    \includegraphics[width=0.99\linewidth]{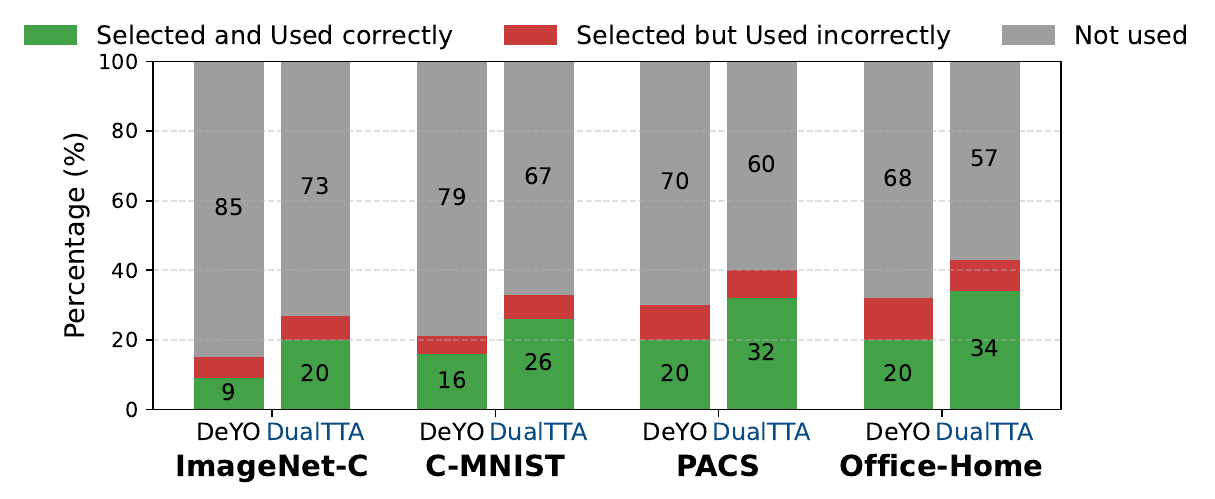}
    \vskip -0.1in
    \caption{Comparison of sample utilization between DualTTA and the prior state-of-the-art method DeYO~\cite{lee2024entropy} across four datasets. DeYO adapts using only a small portion of test samples (green + red), with many of them (red) driving adaptation in the wrong direction. DualTTA, while not perfect, leverages a broader set of samples and misuses a much smaller fraction - achieving better data efficiency and more reliable adaptation.}
    \label{fig:sample_limitation2}
    \vspace{-10pt}
\end{figure}

Given the absence of labeled data, TTA typically updates the model’s parameters by optimizing an unsupervised objective defined on selected test samples. The effectiveness of a TTA method depends critically on the choice of the objective function and the selection of samples for adaptation. One of the most widely adopted objectives in TTA is entropy minimization \cite{wang2021tent, shannon2001mathematical}, as low entropy correlates with the decisiveness of the model. A notable approach, TENT~\cite{wang2021tent}, adapts the model during testing by minimizing the entropy of its predictions on each test sample, encouraging the model to make more confident predictions. However, subsequent studies \cite{lee2024entropy, niu2022efficient} have shown that not all test samples are suitable for adaptation, and incorporating inappropriate samples can lead to model degradation. Motivated by this, several TTA methods have been proposed to focus on selecting the right samples for adaptation \cite{lee2024entropy, wang2021tent, niu2023towards, niu2022efficient}. For instance, EATA \cite{niu2022efficient} and SAR \cite{niu2023towards} retain only test samples with low prediction entropy. More recently, DeYO \cite{lee2024entropy} demonstrated that entropy alone is insufficient to filter out unsuitable samples and proposed an additional criterion based on probability differences.

Existing TTA methods typically adapt using only a small subset of test samples deemed ``reliable'' based on low-entropy predictions, discarding most of the available data. This over-selective strategy suffers from two key failure modes: (i) {\it insufficient utilization}: the coverage is limited, as the pool of confident samples is small under significant distribution shifts; and (ii) {\it misguided adaptation}: confidence does not imply correctness, so high-confidence errors may be misclassified as reliable and, when used for adaptation, can reinforce spurious predictions. For example, DeYO~\cite{lee2024entropy}, a state-of-the-art TTA method, adapts on only $\approx$14\% of test samples (see \Fref{fig:sample_limitation2}), of which only 60-70\% yield beneficial updates (about 9\%). Relying solely on entropy minimization as the adaptation objective is thus both insufficient in scope and misguided in execution.

To address these limitations, we propose \textbf{DualTTA}, a new TTA framework that expands the effective adaptation set through a more principled partitioning of test samples into \textit{likely-correct} and \textit{likely-incorrect} groups. This partitioning is guided by a novel reliability criterion that combines prediction entropy with stability under both semantic-preserving and semantic-altering transformations. By going beyond raw confidence, this criterion better distinguishes informative samples from overconfident outliers, enabling more effective and robust adaptation.




Our paper introduces two core technical contributions. First, we propose a novel reliability criterion based on prediction stability under semantic-preserving and semantic-altering transformations. This criterion enables us to distinguish \textit{likely-correct} samples - those with stable predictions under superficial changes but unstable under content changes - from \textit{likely-incorrect} ones. Based on this classification, we maintain separate adaptation strategies for the two groups. Second, we introduce a dual-objective optimization for test-time adaptation: entropy minimization for \textit{likely-correct} samples to reinforce accurate predictions, and entropy maximization for \textit{likely-incorrect} ones to mitigate overconfidence and unlearn spurious patterns. 

We further provide theoretical analysis showing that our stability-based criterion, combined with the dual-objective formulation, yields a sharper separation between reliable and unreliable samples and leads to better adaptation. Details of this proof will be presented fully in the Supplementary material section.

\section{Related work}
\label{sec:related}

\subsection{Test-Time Adaptation}

TTA aims to enhance model performance in online settings by adapting to unlabeled test data without relying on original training data. Existing methods~\cite{zhang2022memo, lee2023towards, lim2022ttn, niu2023towards, park2023label, wang2021tent, wang2023feature, m_Zhang-Hoai-CVPR23, m_Zhang-etal-BMVC24} mainly follow two approaches: (1) entropy minimization and (2) pseudo-label generation.

\myheading{Entropy minimization.} These methods assume that confident model predictions (low entropy) indicate better performance, and adapt the model using an unsupervised entropy-based loss on selected test samples. TENT \cite{wang2021tent} initiated this line, with follow-ups like EATA \cite{niu2022efficient}, which filtered high-entropy samples to avoid noisy gradients, and SAR \cite{niu2023towards}, which minimized both entropy and loss sharpness using the SAM optimizer \cite{foret2020sharpness} to address small batch sizes and label imbalance. DeYO \cite{lee2024entropy} further identified that low-entropy samples could still lack discriminability and proposed filtering such samples. While effective, these methods struggle to fully leverage the available target data.

\myheading{Pseudo-label generation} derives pseudo-labels for selected samples and uses them in adaptation loss. CoTTA \cite{wang2022continual} generated robust pseudo-labels via augmentation and multiple inferences. FATA \cite{cho2025feature} extended entropy-minimization methods by transforming feature-level representations of selected reliable samples. Though this approach improves sample utilization, it incurs high computational cost, limiting real-world applicability.

\subsection{Prior use of transformation techniques}

Several TTA methods incorporate semantic-altering perturbations, such as spatial transformations, into self-training frameworks \cite{wang2022continual}, often assuming all augmented data is beneficial. This can lead to degraded performance when misleading samples are included. In contrast, our approach selectively identifies which perturbed samples to adapt and how. We use patch-shuffling from DeYO \cite{lee2024entropy} to disrupt spatial structure while preserving texture, encouraging reliance on semantic features. Other spatial perturbations (e.g., cropping, jigsaw, deformation, adversarial augmentation \cite{yun2019cutmix, cubuk2020randaugment}) also help challenge spatial sensitivity.

Semantic-preserving transformations alter low-level features (e.g., color, texture) without changing class semantics, simulating domain shifts. Techniques such as AdaIN~\cite{huang2017arbitrary}, FDA \cite{yang2020fda}, and others \cite{gatys2016image, jackson2019style} support style adaptation. While past TTA work used such perturbations to enforce prediction consistency \cite{pmlr-v119-sun20b, zhang2022memo}, they often overlook harmful cases. Our method instead uses these perturbations to assess prediction stability, improving sample selection and adaptation.

Unlike prior work that uses transformations for regularization, we employ them in a dual-criterion framework for selecting and adapting samples, leading to more reliable and data-efficient test-time adaptation.
\section{Proposed Method}
In this section, we introduce DualTTA, our proposed method.
We first outline the preliminaries of entropy minimization and discuss the limitations of existing sample selection strategies used in entropy-based TTA methods.

\subsection{Limitations of existing entropy-based methods}

TTA considers the setting where a model $f_{\theta}$, parameterized by $\theta$, has been pre-trained on a source dataset 
$\mathcal{D}^{\text{tr}} = \{(\mathbf{x}_i^{\text{tr}}, y_i^{\text{tr}})\}_{i=1}^{N^{\text{tr}}}$
but both the training dataset $\mathcal{D}^{\text{tr}}$ and the labels $y^{\text{tst}}$ of the test data  
$\mathcal{D}^{\text{tst}} = \{\mathbf{x}_i^{\text{tst}}\}_{i=1}^{N^{\text{tst}}}$
are unavailable during inference. Thus TTA methods must rely on unsupervised learning signals, with Shannon entropy minimization \cite{shannon2001mathematical} being a popular approach. This approach optimizes the model to produce low-entropy predictions on selected test samples. Let $\hat{y}_c = f_{\theta}(c |  \mathbf{x})$ be the probability of a sample $\mathbf{x}$ belonging to class $c$. The model's prediction  uncertainty for $\mathbf{x}$ can be measured based on entropy: 
$\text{Ent}_{\theta}(\mathbf{x}) = - \sum_{c \in \mathcal{C}} \hat{y}_c \log \hat{y}_c$,   
where $\mathcal{C}$ denotes the set of class labels.  
Entropy-based TTA methods perform adaptation by optimizing:

\begin{equation}
\mymins{\theta} \sum_{\x \in \mS} \text{Ent}_{\theta}(\mathbf{x}), \label{eq:entropy}
\end{equation}
where $\mathcal{S}$ is a subset of data encountered during test time. The performance of existing entropy-based TTA methods often depends on the choice of $\mathcal{S}$, and these methods have several limitations, as outlined below.

\noindent



\myheading{Suboptimality of the entropy criterion}. 
Recent methods such as EATA~\cite{niu2022efficient}, SAR~\cite{niu2023towards} aim to improve adaptation quality by selecting only confident samples with low entropy: $\mathcal{S} = \{\mathbf{x} \in \mD^{tst} \mid \text{Ent}_{\theta}(\mathbf{x}) < \tau_{\text{ent}} \}$,  where $\tau_{\text{ent}}$ is a predefined entropy threshold. DeYO~\cite{lee2024entropy} further refines sample selection by incorporating structural cues. DeYO employs a sample selection strategy that reduces sensitivity to background information by applying content modifications through the patch-shuffling method. However, some samples depend on fine details or texture-related features for prediction (e.g., in domain-shift datasets like PACS or Office-Home). As a result, this approach may fail to filter out unreliable samples.

To empirically analyze this issue, we conducted an experiment to evaluate the quality of the samples selected for adaptation by comparing their ground-truth labels with the model's predicted labels, i.e., the ``assumed'' labels used by entropy-minimization methods during adaptation. The results in \Fref{fig:sample_limitation2} show that only about 60-70\% of the selected samples have predictions that match the ground truth. For example, in the PACS dataset, although 28\% of test samples were used for adaptation, only 15\% were correctly predicted. Including incorrectly predicted samples in the adaptation process, and further minimizing prediction entropy on them, can degrade performance.

\myheading{Dependence on Entropy Minimization.}
Based on the idea of entropy minimization, current TTA methods devote significant effort to select a small ``high-quality'' subset of test samples and then apply entropy minimization exclusively on this set. While low-entropy samples often constitute around 25-30\% of ImageNet-C samples, methods like DeYO utilize only 14\%, discarding the remaining potentially useful data (see \Fref{fig:sample_limitation2}). Such aggressive filtering, though intended to ensure reliable updates, drastically reduces the number of samples available for adaptation. This sparse coverage restricts the model's exposure to diverse target-domain features particularly detrimental for underrepresented classes, and ultimately undermines the statistical strength and robustness of the adaptation process.

Addressing these limitations, we introduce a double-criterion selection mechanism based on consistency under semantic-preserving and semantic-altering transformations. This approach filters out samples whose predictions remain stable under content changes but vary with superficial changes. By combining these two transformation-based criteria, our strategy improves the reliability of selected samples beyond what raw entropy scores provide, while also recovering a significant portion of previously discarded data, leading to higher data efficiency. 
\subsection{Overview of DualTTA}
\label{TTAoverview}
\begin{figure*}[t]
    \centering
    \includegraphics[width=0.95\linewidth]{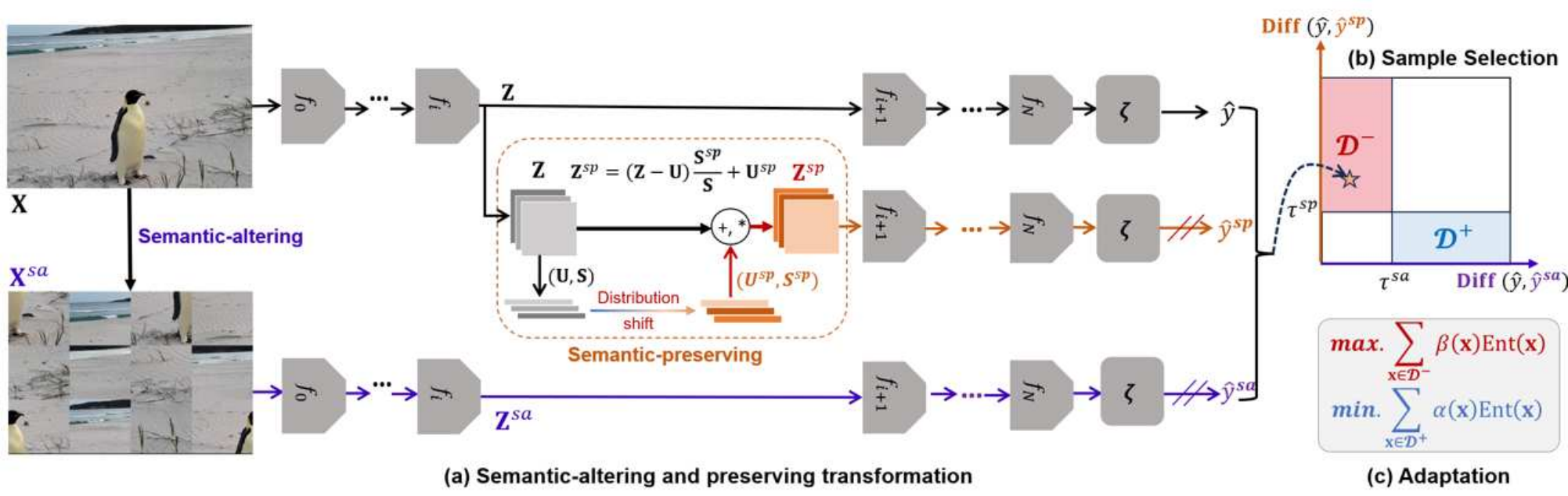}
    \caption{\textbf{Overview of DualTTA.} DualTTA apply 2 transformations: semantic-altering and semantic-preserving on each sample. The model's predictions for the original samples and its transformed variants are compared to determine reliability. Samples with stable predictions under semantic alteration but varying predictions under semantic preservation are classified as \textbf{\textcolor[HTML]{B22222}{likely-incorrect}}, while those with unstable predictions under content-altering but stable under content-preserving are determined as \textbf{\textcolor[HTML]{2A52BE}{likely-correct}}. likely-correct samples undergo entropy minimization, while likely-incorrect ones are penalized via entropy maximization to prevent adaptation degradation. 
    }
    \label{fig:overview}
    \vspace{-10pt}
\end{figure*}

\Fref{fig:overview} illustrates the overall pipeline of DualTTA. It uses two transformation functions, simulating \textbf{semantic-altering} and \textbf{semantic-preserving}, to generate pseudo-labels, which are then compared with the original predictions to guide sample selection. Semantic-altering, inspired by \cite{lee2024entropy}, involves dividing images into patches and shuffling them to disrupt content structure, while semantic-preserving operates in the latent space by only changing the mean and standard deviation of the distribution without affecting the semantic content. These transformations serve distinct purposes in generating pseudo-labels, which are subsequently compared with the original predictions. The differences between the predictions serve as a metric for selecting and processing samples during adaptation.

Samples whose prediction outputs remain stable under semantic-preserving transformations but exhibit significant variations under semantic-altering transformations are considered \textbf{likely-correct} and are denoted as $\mathcal{D}^+$. Conversely, samples whose prediction outputs change dramatically under semantic-preserving transformations while being less affected by semantic-altering transformations are considered \textbf{likely-incorrect} and are denoted as $\mathcal{D}^-$. DualTTA performs model adaptation by minimizing the entropy of the model's predictions for likely-correct samples $\mathcal{D}^+$ while maximizing the entropy of the model's predictions for likely-incorrect ones $\mathcal{D}^-$.

The rationale behind DualTTA is to adapt models more effectively by boosting confidence in likely-correct predictions while suppressing overconfidence in likely-incorrect ones. It distinguishes between these cases by analyzing prediction behavior under semantic-preserving and semantic-altering transformations - reinforcing stable, semantically consistent predictions and penalizing unstable, potentially spurious ones. Unlike other TTA methods that rely solely on entropy minimization applied uniformly to low-entropy (i.e., confident) samples, DualTTA introduces a dual adaptation strategy: it explicitly selects separate sets of likely-correct and likely-incorrect test samples and treats them differently. This design not only moves beyond simplistic confidence-based selection by incorporating transformation-driven reliability but also mitigates the harm caused by overconfident mispredictions during adaptation.

\subsection{Likely-correct and -incorrect Determination}


We now describe our novel sample selection strategy, which evaluates prediction stability before and after applying input transformations. For a given input sample $\mathbf{x}$, we assume the existence of a \textit{semantic-preserving} transformation yielding $\mathbf{x}^{sp}$ and a \textit{semantic-altering} transformation yielding $\mathbf{x}^{sa}$. We will discuss such transformations in \Sref{sec:transformation}. Let $\hat{y}$, $\hat{y}^{sp}$, and $\hat{y}^{sa}$ denote the predicted probability vectors for the original input $\mathbf{x}$, its semantic-preserving variant, and its semantic-altering variant, respectively.

We define the difference between two probability vectors $\hat{y}$ and $y$ as: 
\begin{align}
    \textrm{Diff}(\hat{y}, y) = \hat{y}_k - y_k, \textrm{where } k = \myargmax{c} \hat{y}_c.
\end{align}


From the test set $\mD^{tst}$, we create two subsets  $\mD^+, \mD^-$ for the likely-correct and -incorrect samples:
\begin{align}
    \mD^+ = &\{\x \in \mD^{tst}|\text{Diff}(\hat{y},\hat{y}^{sa})>\tau^{sa},\text{Diff}(\hat{y},\hat{y}^{sp})<\tau^{sp}\}, \nonumber\\
    \mD^- = &\{\x \in \mD^{tst}|\text{Diff}(\hat{y},\hat{y}^{sa})<\tau^{sa},\text{Diff}(\hat{y},\hat{y}^{sp})>\tau^{sp}\}, \nonumber
\end{align}
where $\tau^{sa}$ and $\tau^{sp}$ are pre-defined thresholds. The subset $\mathcal{D}^+$ comprises samples with stable predictions under semantic-preservation but significant changes when structural content is perturbed. In contrast, the subset $\mathcal{D}^-$ includes samples that are highly sensitive to superficial changes while exhibiting minimal prediction variations when structural content is altered.


\subsection{Loss function for adaptation}

For TTA, we propose to minimize the following loss:
\begin{align}
&\mL_{Dual} = \mL^+ - \lambda \mL^-, \textrm{where } \\
 & \mL^+ = \sum_{\x \in \mD^+}{\alpha(\x)\text{Ent}(\x)}, \textrm{and }  \mL^- = \sum_{\x \in \mD^-}\beta(\x)\text{Ent}(\x). \nonumber
\end{align}

In the above, $\lambda$ is the trade-off coefficient between the two loss terms, and $\alpha(\x)$ and $\beta(\x)$ are samples weight that reflects the model's confidence in the sample, as well as the degree of alignment/misalignment between the model and the sample $\x$ under semantic-altering and semantic-preserving transformations. Specifically:
\begin{align}
&\alpha(\x) = e^{\text{Ent}_0 - \text{Ent}(\x)}+e^{\text{Diff}(\hat{y}, \hat{y}^{sa})} +e^{\text{Diff}_0 - \text{Diff}(\hat{y},\hat{y}^{sp})}, \\
&\beta(\x)=e^{\text{Ent}_0-\text{Ent}(\x)},
\end{align}
where $\text{Ent}_0,\text{Diff}_0$ are pre-defined normalization factors. The sample weight $\alpha(\mathbf{x})$ is higher when the model is confident in its prediction (low entropy), the difference between the model's predictions for the original and content-altered samples is large, and the difference between the predictions for the original and content-preserved samples is small. Then, $\theta$ will be updated based on $\mathcal{L}_{Dual}$ for adaptation: $\theta = \theta - \nabla_\theta \mathcal{L}_{Dual}$.


\subsection{Image Transformations \label{sec:transformation}}


Our DualTTA framework relies on semantic-preserving and semantic-altering transformations, but it is not limited to any specific ones, such transformations are generally easy to construct. For image data, operations that significantly disrupt spatial arrangement (e.g., shuffling image patches) typically alter the object category and serve as semantic-altering transformations. In contrast, modifications that slightly adjust pixel intensities or make minor spatial changes without altering the overall structure are considered semantic-preserving.

Importantly, these transformations can be applied not only at the input level but also at intermediate stages of the model (e.g., feature maps at one of the first layers of backbone), offering flexibility in their design. In this paper, we adopt patch shuffling, following DeYO~\cite{lee2024entropy}, as our semantic-altering transformation, which disrupts spatial structure while preserving texture, thereby encouraging reliance on high-level semantic features and allowing fair comparison with DeYO. In the remainder of this section, we describe the specific semantic-preserving transformation used in our experiments, while noting that other variants are equally applicable.

Given a batch of test samples $\X \in \mathbb{R}^{B{\times}C{\times}H{\times}W}$ and a pre-trained model $f_{\theta}$ composed of $N$ layers $f_1, f_2,...,f_N$. The feature map $\Z_i \in \mathbb{R}^{B{\times}C_i{\times}H_i{\times}W_i}$ at the output of the first $i$ layers is obtained by sequentially applying these layers to~$\X$: $\Z_i = f_i \circ f_{i-1} \circ \dots \circ f_1 (\X)$. We denote $\U \in \mathbb{R}^{B{\times}C_i}$ and $\S \in \mathbb{R}^{B{\times}C_i}$ as the channel-wise mean and standard deviation of each instance in the batch, respectively, which is defined as follows:
\begin{align}
\nonumber
&\U(b,c) = \frac{1}{H_iW_i}\sum_{h=1}^{H_i}\sum_{w=1}^{W_i}\Z(b,c,h,w),\\
\nonumber
&\S(b,c) = \sqrt{\frac{1}{H_iW_i}\sum_{h=1}^{H_i}\sum_{w=1}^{W_i}(\Z(b,c,h,w)-\U(b,c))^2}.
\end{align}

From an abstract perspective, feature statistics encapsulate key characteristics of a given domain, such as color, texture, and contrast, often referred to as style statistics in previous studies \cite{huang2017arbitrary, li2021on}. In out-of-distribution scenarios, these statistics frequently diverge from those in the training data due to inherent domain differences \cite{wang2019transferable, gao2021representative}.

We simulate the distribution shift process while maintaining the semantic information to identify samples within the batch that are sensitive to domain shifts. Specifically, this is achieved by modifying their style statistics from $(\U, \S)$ to $(\U^{sp}, \S^{sp})$ according to the following formula, where the subscript $i$ is omitted for brevity.
\begin{align}
    \U^{sp}  = \U + \epsilon_\U\U^\sigma, \textrm{and }
    \S^{sp} = \S + \epsilon_\S \S^\sigma, 
\end{align}
where $\epsilon_\U,\epsilon_\S \in \mathbb{R}^{B{\times}1}$ are random Gaussian noises, and $\U^\sigma,\S^\sigma \in \mathbb{R}^{1{\times}C_i}$ are the standard deviation for the entries in $\U$ and $\S$ across the batch dimension, respectively. More formally, $\U^\sigma$ which is defined as: 
\begin{align}
        \U^{\sigma}(c) &=\sqrt{\frac{1}{B}\sum_{b=1}^B\left(\U(b,c) - \frac{1}{B}\sum_{b=1}^B \U(b,c)\right)^2}.
\end{align}
The matrix $\S^{\sigma}$ is defined similarly. The semantic-preserving transformation is done as follows:
\begin{equation}
\resizebox{\linewidth}{!}{
    $\Z^{sp}(b,c,h,w)= (\Z(b,c,h,w)-\U(b,c))\frac{\S^{sp}(b,c)}{\S(b,c)}+\U^{sp}(b,c). $ \nonumber 
}
\end{equation}
After transformation, the new feature $\Z^{sp}$ is fed into the next layer of the network for making probability predictions $\hat{y}^{sp}\in \mathbb{R}^{B*|\mathcal{C}|}$, where $|\mathcal{C}|$ denotes total number of classes, i.e., $\hat{y}^{sp}= \zeta \circ f_N \circ\dots \circ f_{i+1}(\Z^{sp}).$

\section{Experiments}
We conducted extensive experiments across diverse benchmarks to evaluate the effectiveness, robustness, and data efficiency of DualTTA, comparing it with state-of-the-art test-time adaptation methods and analyzing the contribution of dual strategies and likely-incorrect sample.
\subsection{Settings}
\myheading{Benchmark datasets.}
We conduct experiments on commonly used benchmarks covering three out-of-distribution scenarios: spurious correlation, domain shift, and data corruption. To evaluate model performance under extreme spurious correlation shifts, we use ColoredMNIST and Waterbirds \cite{lee2024entropy}. For domain shift, we test on PACS \cite{li2017deeper} and Office-Home \cite{venkateswara2017deep}.
To measure robustness against data corruption, we employ ImageNet-C \cite{hendrycks2019benchmark}, containing 15 types of corruption with five levels of severity.
\\
\myheading{Backbone and normalization.}
Experiments are conducted using architectures that integrate three distinct normalization techniques: Batch Normalization (BN), Group Normalization (GN), and Layer Normalization (LN). Specifically, ResNet-18 and ResNet-50 are employed for BN and GN, while the ViT-Base model was utilized for LN. For details, we utilize ResNet18-BN for ColoredMNIST, ResNet50-BN for Waterbirds, PACS, Office-Home and ResNet50-GN, ViT-B-LN for ImageNet-C. 

\myheading{Test scenarios.} For all datasets, we follow the mild scenario proposed by \cite{wang2021tent}. For ImageNet-C, we also experiment with two additional test scenarios suggested by \cite{niu2023towards} with imbalanced label shift and mixed distribution, under the highest level of corruption (Level 5).

\myheading{Pretrained models.} For ImageNet-C, all models are pretrained on the ImageNet dataset and subsequently adapted during the testing phase. For the PACS, Office-Home datasets, the models are initially pre-trained on a source domain and later adapted to other target domains at test time. 

\myheading{Baseline and hyper-parameters.} We evaluate the performance of DualTTA against state-of-the-art TTA methods, including TENT~\cite{wang2021tent}, SAR~\cite{niu2023towards}, EATA~\cite{niu2022efficient}, and DeYO~\cite{lee2024entropy}, using a consistent batch size of 64 across all experiments. For ResNet-18 and ResNet-50, we set the learning rate to $0.5{\times}10^{-3}$, while for ViT-B, we use $10^{-4}$. 

For DualTTA, we set the semantic-altering threshold to $\tau^{sa}=0.4$ and define the semantic-preserving threshold as $\tau^{sp}=0.7$, the difference normalization factor $\text{Diff}_0=0.7$, and the trade-off coefficient $\lambda=0.5$. Additionally, we apply semantic-preserving transformations at layer $i=1$ for ResNet and $i=7$ for ViT-B. Further explanation will be presented in the Appendix.
\subsection{Comparison with the state-of-the-art}
\subsubsection{Performance under spurious correlation shifts.}


The proposed method, DualTTA, demonstrates superior performance on datasets exhibiting spurious correlations between objects and backgrounds. Since the spurious correlation dataset contains $\mathcal{C}$ labels based on both background and object, it is divided into $2\mathcal{C}$ groups. Avg Acc and Worst Acc represent the average accuracy and the lowest accuracy across these $2\mathcal{C}$ groups, respectively. Specifically, \Tref{tab:combined} shows that DualTTA achieves an average accuracy improvement of 4.54\% and 1.37\% over the second-best method, DeYO, on the ColoredMNIST and Waterbirds datasets. 

\subsubsection{Performance under domain shift.}

\Tref{tab:combined} presents the average accuracy when the model is pre-trained on one of the four domains and then used for inference on the remaining three domains in the Office-Home and PACS dataset. 

For the Office-Home dataset consisting of four domains: Art, Clipart, Product, and Real-world, performance improves by an average of 2.43\% compared to DeYO. See Supplementary material for more details, where performance improvements reach up to 6.02\% when the model is trained on Art and tested with TTA on Clipart.

On the PACS dataset comprising four domains: Art, Cartoon, Photo, and Sketch performance improves by an average of 0.96\% compared to the second-best method, DeYO. Supplementary material provides more details, showing that performance gains can reach up to 7.02\% when the model is trained on Photo and tested with TTA on Sketch.  

\begin{table}[t]
    \centering
    \caption{Performance of DualTTA and baselines across several benchmarks: ColoredMNIST, Waterbirds, Office-Home, PACS. 
    }
    \resizebox{1.05\linewidth}{!}{
    \begin{tabular}{l||cc|cc|c|c|c}
    \toprule
    \multirow{2}{*}{\textbf{Methods}}
      & \multicolumn{2}{c|}{\textbf{ColoredMNIST}}
      & \multicolumn{2}{c|}{\textbf{Waterbirds}}
      & \multirow{2}{*}{\textbf{Office-Home}}
      & \multirow{2}{*}{\textbf{PACS}}
      \\
    \cmidrule(lr){2-3} \cmidrule(lr){4-5}
      & Avg Acc & Worst Acc
      & Avg Acc & Worst Acc
      &  &  &  \\
    \midrule
    No adapt
      & 63.46 & 20.14
      & 81.81 & 62.90
      & 59.14
      & 72.34
      \\ 
    Tent
      & 56.51 &  8.99
      & 83.63 & 54.99
      & 61.08
      & 74.96
      \\ 
    SAR
      & 58.10 & 11.82
      & 82.91 & 53.59
      & 59.82
      & 73.26
      \\ 
    EATA
      & 60.65 & 17.59
      & 80.65 & 52.18
      & 52.34
      & 72.46
      \\ 
    DeYO
      & 77.98 & 65.59
      & 87.17 & 71.98
      & 59.08
      & 75.16
      \\ 
    \rowcolor[HTML]{ACE5EE}
    \textbf{DualTTA}
      & \textcolor{red}{\textbf{82.12}} & \textcolor{red}{\textbf{68.82}}
      & \textcolor{red}{\textbf{88.44}} & \textcolor{red}{\textbf{72.53}}
      & \textcolor{red}{\textbf{61.51}}
      & \textcolor{red}{\textbf{76.02}}
      \\
    \bottomrule
    \end{tabular}
    }
    \label{tab:combined}
\end{table}

\subsubsection{Performance under corruption shift.}
\begin{table*}[t]
    \centering
    \captionof{table}{Model performance on ImageNet-C with corruption level 5. The best results are colored {\color[HTML]{FE0000} {\textbf{bold red}}}. DualTTA achieves the best performance on most corruption types, showing strong robustness under severe distribution shifts with different normalization layers.}
    
    \resizebox{1\textwidth}{!}{
    \begin{tabular}{l||ccc|cccc|cccc|cccc||l}
    \toprule
    \multicolumn{1}{c||}{} & \multicolumn{3}{c|}{Noise} & \multicolumn{4}{c|}{Blur} & \multicolumn{4}{c|}{Weather} & \multicolumn{4}{c||}{Digital} &  \\ 
    \cmidrule{2-16}
    \multicolumn{1}{c||}{\multirow{-2}{*}{Methods}} & Gauss. & Shot & Impul. & Defoc. & Glass & Motion & Zoom & Snow & Frost & Fog & Bright. & Contr. & Elastic & Pixel & JPEG & \multirow{-2}{*}{\textbf{Avg}} \\ 
    \midrule
    ResNet50-BN & 0.30 & 0.37 & 0.35 & 0.24 & 0.21 & 0.43 & 0.66 & 1.06 & 1.30 & 1.67 & 2.29 & 0.86 & 0.66 & 0.73 & 0.84 & 0.81 \\
    + Tent & 18.32 & 21.70 & 18.54 & 18.71 & 18.57 & 33.58 & 44.91 & 45.74 & 46.09 & 61.78 & 70.54 & 31.67 & 48.35 & 51.39 & 51.55 & 38.76 \\
    + SAR & 16.60 & 20.56 & 19.93 & 13.74 & 12.86 & 34.01 & 45.33 & 46.11 & 45.66 & 61.78 & 70.34 & 32.36 & 48.43 & 52.58 & 48.92 & 37.95 \\
    + EATA & {\color[HTML]{FE0000} \textbf{24.96}} & {\color[HTML]{FE0000} \textbf{28.51}} & 26.43 & 20.28 & 21.84 & 36.95 & 46.84 & 48.27 & 47.91 & 63.18 & 70.47 & 35.74 & 51.12 & 53.47 & 52.09 & 41.87 \\
    + DeYO & 17.09 & 25.72 & 26.74 & 19.21 & 24.14 & 33.70 & 44.21 & 45.97 & 48.02 & 59.84 & 69.86 & {\color[HTML]{FE0000} \textbf{41.25}} & 51.90 & 57.14 & 54.69 & 41.30 \\ 
    \rowcolor[HTML]{ACE5EE} 
    \rowcolor[HTML]{ACE5EE} 
    \textbf{+ DualTTA} & $23.69$ & $24.38$ & {\color[HTML]{FE0000} $\textbf{29.59}$} & {\color[HTML]{FE0000} $\textbf{20.30}$} & {\color[HTML]{FE0000} $\textbf{25.97}$} & {\color[HTML]{FE0000} $\textbf{42.33}$} & {\color[HTML]{FE0000} $\textbf{51.13}$} & {\color[HTML]{FE0000} $\textbf{54.03}$} & {\color[HTML]{FE0000} $\textbf{50.76}$} & {\color[HTML]{FE0000} $\textbf{65.77}$} & {\color[HTML]{FE0000} $\textbf{71.90}$} & $34.81$ & {\color[HTML]{FE0000} $\textbf{56.64}$} & {\color[HTML]{FE0000} $\textbf{58.79}$} & {\color[HTML]{FE0000} $\textbf{57.77}$} & {\color[HTML]{FE0000} $\textbf{44.52}$} \\
    \midrule
    ResNet50-GN & 22.09 & 23.03 & 22.04 & 19.79 & 11.40 & 21.46 & 25.04 & 40.28 & 46.97 & 34.02 & 68.81 & 36.25 & 18.51 & 29.24 & 52.60 & 31.44 \\
    + Tent & 8.91 & 9.57 & 8.71 & 9.22 & 7.24 & 12.08 & 14.75 & 12.80 & 13.36 & 1.31 & 69.90 & 40.27 & 3.37 & 49.51 & 52.45 & 20.90 \\
    + SAR & {\color[HTML]{FE0000} \textbf{39.96}} & 42.08 & 41.35 & 19.35 & 22.01 & 37.65 & 39.12 & 24.89 & 46.87 & 54.32 & 72.37 & 49.37 & 5.82 & 54.89 & 57.31 & 40.49 \\
    + EATA & 37.42 & 41.49 & 39.64 & {\color[HTML]{FE0000} \textbf{29.70}} & {\color[HTML]{FE0000} \textbf{27.01}} & 37.90 & {\color[HTML]{FE0000} \textbf{41.82}} & {\color[HTML]{FE0000} \textbf{51.54}} & {\color[HTML]{FE0000} \textbf{47.76}} & {\color[HTML]{FE0000} \textbf{58.30}} & 71.64 & 51.66 & 26.76 & 59.18 & 58.88 & {\color[HTML]{FE0000} \textbf{45.38}} \\
    + DeYO & 36.00 & 46.13 & {\color[HTML]{FE0000} \textbf{46.99}} & 14.73 & 20.42 & 12.58 & 17.41 & 27.09 & 26.98 & 33.42 & 66.33 & 41.98 & {\color[HTML]{FE0000} \textbf{22.56}} & 44.01 & 50.70 & 33.82 \\ 
    \rowcolor[HTML]{ACE5EE} 
    \textbf{+ DualTTA} & $25.96$ & {\color[HTML]{FE0000} $\textbf{46.31}$} & $44.43$ & $24.81$ & $26.12$ & {\color[HTML]{FE0000} $\textbf{42.06}$} & $18.92$ & $23.27$ & $25.75$ & $20.73$ & {\color[HTML]{FE0000} $\textbf{72.96}$} & {\color[HTML]{FE0000} $\textbf{53.80}$} & $21.55$ & {\color[HTML]{FE0000} $\textbf{61.99}$} & {\color[HTML]{FE0000} $\textbf{60.17}$} & $41.83$ \\
    \midrule
    ViTBase-LN & 35.09 & 32.16 & 35.87 & 31.42 & 25.31 & 39.45 & 31.55 & 24.47 & 30.13 & 54.74 & 64.48 & 48.98 & 34.20 & 53.17 & 56.45 & 39.83 \\
    + Tent & 52.58 & 51.58 & 53.55 & 52.75 & 47.82 & 56.47 & 48.01 & 10.00 & 31.78 & 67.38 & 74.28 & 67.19 & 50.79 & 66.75 & 64.73 & 53.04 \\
    + SAR & 51.92 & 51.41 & 52.89 & 51.62 & 48.59 & 55.45 & 49.61 & 12.84 & 49.90 & 66.79 & 73.03 & 65.62 & 52.62 & 63.90 & 63.15 & 53.96 \\
    + EATA & {\color[HTML]{FE0000} \textbf{55.87}} & {\color[HTML]{FE0000} \textbf{56.14}} & {\color[HTML]{FE0000} \textbf{56.96}} & {\color[HTML]{FE0000} \textbf{57.36}} & 53.28 & 62.00 & {\color[HTML]{FE0000} \textbf{58.63}} & 61.99 & 59.87 & 71.47 & 75.54 & {\color[HTML]{FE0000} \textbf{68.66}} & 63.23 & 69.20 & 66.57 & 62.44 \\
    + DeYO & 49.77 & 53.27 & 54.71 & 48.79 & 50.17 & 55.87 & 51.82 & 57.64 & 61.09 & 64.78 & 75.67 & 62.34 & 58.48 & 67.41 & 67.12 & 58.59 \\ 
    \rowcolor[HTML]{ACE5EE} 
    \textbf{+ DualTTA} & $54.45$ & $55.32$ & $55.47$ & $55.74$ & {\color[HTML]{FE0000} $\textbf{54.54}$} & {\color[HTML]{FE0000} $\textbf{62.01}$} & $57.48$ & {\color[HTML]{FE0000} $\textbf{62.58}$} & {\color[HTML]{FE0000} $\textbf{67.39}$} & {\color[HTML]{FE0000} $\textbf{71.29}$} & {\color[HTML]{FE0000} $\textbf{76.85}$} & $67.55$ & {\color[HTML]{FE0000} $\textbf{64.78}$} & {\color[HTML]{FE0000} $\textbf{71.32}$} & {\color[HTML]{FE0000} $\textbf{68.05}$} & {\color[HTML]{FE0000} $\textbf{63.00}$} \\

    \bottomrule
    \end{tabular}%
    }
    \label{tab:image-net5}
    \vspace{-1em}
\end{table*}

To ensure reproducibility and assess stability, we run all experiments three times using a fixed random seed of 2024 and report the mean and standard deviation of the resulting accuracies.
\Tref{tab:image-net5} presents the accuracy when the model is pre-trained on ImageNet and tested on 15 types of level-5 corruptions using different backbone architectures: ResNet50-BN, ResNet50-GN and ViT-B-LN. DualTTA achieves improvements of up to \textbf{8.06\%} and 2.65\% in average over the second-best method, EATA, on ResNet-50-BN. On ViT-B-LN, DualTTA improves up to 7.52\% and 0.56\% in average over DeYO .


To better reflect challenging test conditions in real-world deployment, SAR \cite{niu2023towards} introduces two more realistic test protocols: (i) dynamic changes in the ground-truth label distribution during testing, resulting in label imbalance across different corruptions; and (ii) the presence of concurrent distribution shifts.
The results in Table \ref{tab:mixed_shift} show that DualTTA consistently achieves strong performance across all settings and architectures. Under the imbalanced label condition, DualTTA outperforms the second-best baseline SAR by 1.10\% on ViTBase-LN. In mixed shift setting, DualTTA significantly outperforms prior methods: 1.41\% on ResNet-BN vs SAR, and 1.17\% on ViTBase-LN vs DeYO.

\subsection{Ablation Studies}

\subsubsection{Dual-transformation sample selection.}
To understand the importance of combining semantic-altering and semantic-preserving transformations, we conduct an experiment comparing the performance of DualTTA with variations that exclude one of these transformations as a criterion for filtering samples. The experiment is conducted on three datasets: ColoredMNIST, PACS, and OfficeHome. As shown in \Tref{tab:impact_filtering}, the semantic-altering condition plays a crucial role in ColoredMNIST, while the semantic-preserving transformation significantly enhances performance on PACS and OfficeHome. In more detail, applying the semantic-altering condition improves performance by 8.49\% and 4.19\% on ColoredMNIST and Waterbirds, respectively. Similarly, using semantic-preserving condition improves performance by 3.05\% on ColoredMNIST, 1.64\% on average across PACS, and 2.81\% on average across OfficeHome. 

\begin{figure}[t]
\centering
\includegraphics[width=0.8\linewidth]{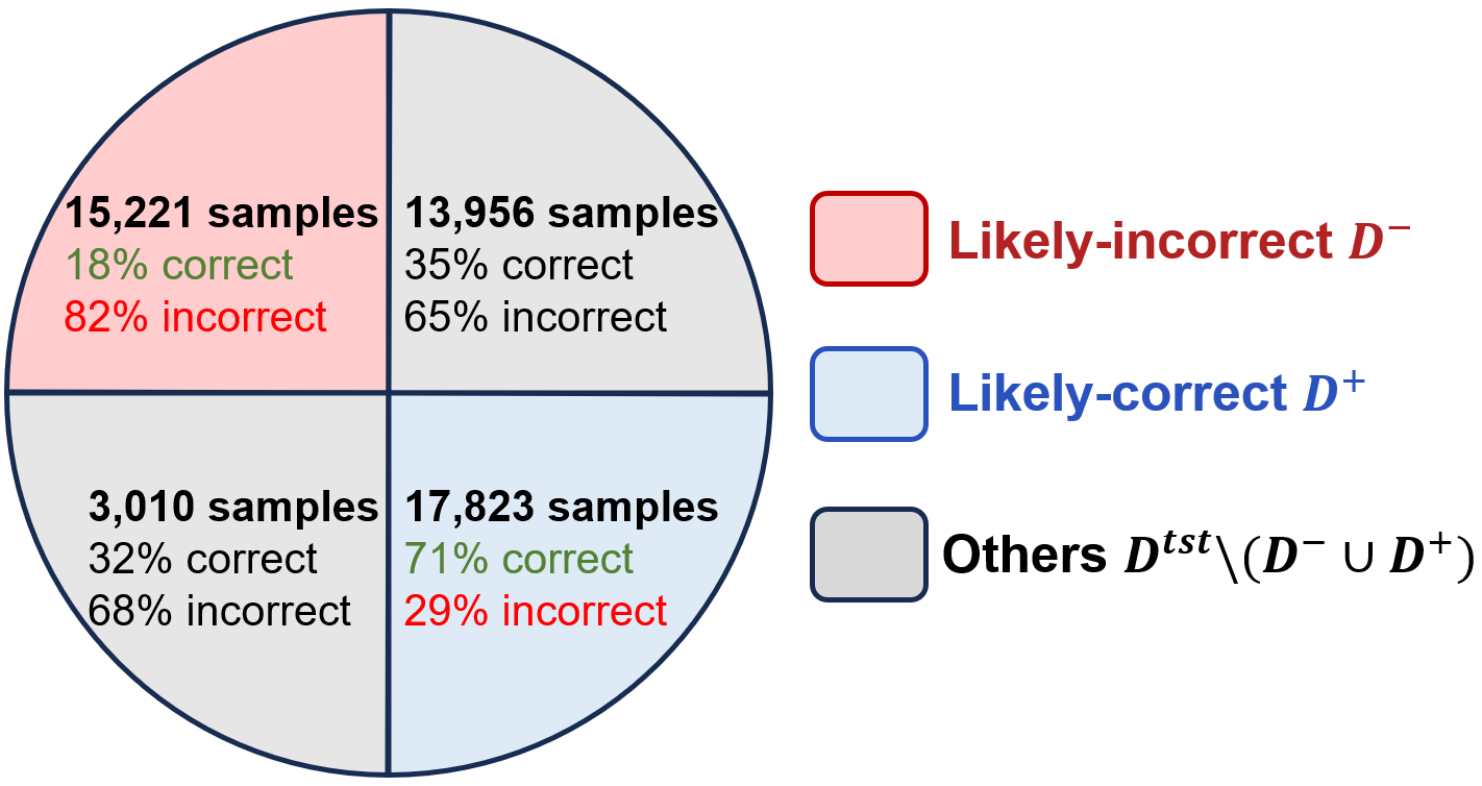}
\vskip -0.1in
\caption{Quadrant-based accuracy analysis of 50000 impulse noise samples from ImageNet-C using our selection strategies. The blue region (\textbf{\textcolor[HTML]{2A52BE}{likely-correct $\mathcal{D}^+$}}) contains a majority of correct samples, while the red region (\textbf{\textcolor[HTML]{B22222}{likely-incorrect $\mathcal{D}^-$}}) is dominated by incorrect predictions, demonstrating that DualTTA effectively separates high- and low-accuracy predictions.}
\label{dual-role}
\end{figure}

We also provide an empirical justification for these 2 selection strategies, by evaluating their accuracy relative to the ground truth (\% correct and \% incorrect) across four quadrants on 50000 samples from ImageNet-C. As illustrated in \Fref{dual-role}, $\mD^+$ is dominated by samples with predictions that correctly match the ground truth, at 71\% vs 29\% incorrect, while $\mD^-$is primarily composed of samples with incorrect predictions, at 82\% vs 18\% correct. 

\begin{table}[t]
    \centering
    \caption{Average accuracy on ImageNet-C at severity level 5 under imbalanced label distribution and mixed distribution. 
    }
    \resizebox{1.05\linewidth}{!}{%
    \begin{tabular}{l||ccc|ccc}
        \toprule
        & \multicolumn{3}{c|}{\textbf{Imbalanced label}} 
        & \multicolumn{3}{c}{\textbf{Mixed shift}} \\ 
        \cmidrule(lr){2-4}\cmidrule(lr){5-7}
        \textbf{Methods} 
          & ResNet-BN & ResNet-GN & ViTBase-LN 
          & ResNet-BN & ResNet-GN & ViTBase-LN \\ 
        \midrule
        No adapt    & 32.11 & 31.44 & 39.83 & 0.14  & 31.44 & 54.37 \\
        Tent        & 42.80 & 29.08 & 52.47 & 2.35  & 13.16 & 52.82 \\
        SAR         & 46.27 & {\color[HTML]{FE0000}\textbf{45.67}} & 62.10 & 26.10 & 35.09 & 54.33 \\
        EATA        & {\color[HTML]{FE0000}\textbf{47.14}} & 34.86 & 53.60 & 19.62 & 38.50 & 58.64 \\
        DeYO        & 41.03 & 41.54 & 58.68 & 11.63  & {\color[HTML]{FE0000}\textbf{39.26}} & 69.73 \\ 
        \rowcolor[HTML]{ACE5EE} 
        \textbf{DualTTA}
          & $43.83$ & $38.88$ & {\color[HTML]{FE0000}$\textbf{63.30}$}
          & {\color[HTML]{FE0000}$\textbf{27.51}$} & $38.92$ & {\color[HTML]{FE0000}$\textbf{70.90}$} \\
        \bottomrule
    \end{tabular}
    }
    \label{tab:mixed_shift}
\end{table}

\begin{figure}[t]
\centering
\includegraphics[width=0.98\linewidth]{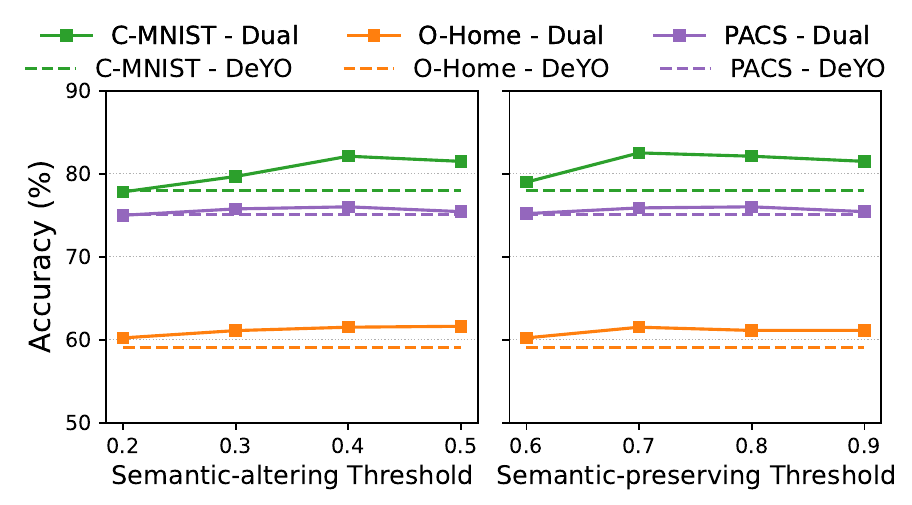}
\vskip -0.1in
\caption{Sensitivity of DualTTA to semantic-altering and semantic-preserving thresholds on C-MNIST, O-Home, and PACS. DualTTA maintains stable performance across a wide range of values, showing low sensitivity to these parameters. DeYO~\cite{lee2024entropy} is shown for reference and is outperformed by DualTTA. \label{hyperparam}
}
\vspace{-1.5em}
\end{figure}

\begin{table}[t]
\centering
\caption{
Impact of dual transformation filtering of DualTTA on different datasets. 
Each column indicates whether semantic-altering or semantic-preserving transformations are used (\ding{51}) or not (\ding{55}). 
Results show that employing both transformations consistently yields the highest accuracy, highlighting the roles of semantic-altering and semantic-preserving filters in DualTTA.
}
\label{tab:impact_filtering}
\resizebox{\columnwidth}{!}{
\begin{tabular}{l|c|c|c}
\toprule
\rowcolor[HTML]{E5E4E2}
\textbf{DualTTA on} & \textbf{Semantic-Altering} & \textbf{Semantic-Preserving} & \textbf{Acc} \\
\midrule
\multirow{3}{*}{ColoredMNIST} 
 & \ding{51} & \ding{51} & \cellcolor[HTML]{ACE5EE}\textbf{82.12} \\
 & \ding{55} & \ding{51} & 74.63 \\
 & \ding{51} & \ding{55} & 79.17 \\
\midrule
\multirow{3}{*}{PACS} 
 & \ding{51} & \ding{51} & \cellcolor[HTML]{ACE5EE}\textbf{76.02} \\
 & \ding{55} & \ding{51} & 75.80 \\
 & \ding{51} & \ding{55} & 74.48 \\
\midrule
\multirow{3}{*}{OfficeHome} 
 & \ding{51} & \ding{51} & \cellcolor[HTML]{ACE5EE}\textbf{61.51} \\
 & \ding{55} & \ding{51} & 61.20 \\
 & \ding{51} & \ding{55} & 59.70 \\
\bottomrule
\end{tabular}
}
\end{table}


\begin{table}[t]
\centering
\caption{
Impact of utilizing likely-incorrect samples on different methods. 
The additional column indicates whether the method incorporates likely-incorrect samples (\ding{51}) or not (\ding{55}). 
Results show that selectively leveraging these samples improves performance across all methods, confirming their complementary role of entropy maximization in test-time adaptation.
}
\label{tab:impact_incorrect}
\resizebox{\columnwidth}{!}{
\begin{tabular}{l|c|c|c|c}
\toprule
\rowcolor[HTML]{E5E4E2}
\textbf{Original Method} & \textbf{Likely-Incorrect} & \textbf{Colored MNIST} & \textbf{PACS} & \textbf{Office-Home} \\
\midrule
\rowcolor[HTML]{ACE5EE}
\textbf{DualTTA} (proposed) & \ding{51} & \textbf{82.12} & \textbf{76.02} & \textbf{61.51} \\
DualTTA (ablated) & \ding{55} & 80.65 & 75.52 & 61.10 \\
\midrule
DeYO (original) & \ding{55} & 77.98 & 75.16 & 59.08 \\
\rowcolor[HTML]{E0FFFF}
DeYO (improved) & \ding{51} & 79.17 & 75.65 & 59.70 \\
\midrule
EATA (original) & \ding{55} & 60.65 & 72.46 & 52.34 \\
\rowcolor[HTML]{E0FFFF}
EATA (improved) & \ding{51} & 62.31 & 73.47 & 55.72 \\
\bottomrule
\end{tabular}
}
\end{table}

\begin{table}[t]
\centering
\caption{Data efficiency analysis. \textit{``\%corr-adapt"} represents the share of likely-correct samples whose predictions match the ground truth and likely-incorrect samples whose predictions differ from the ground truth. \textit{``\% adapt"} represents the share of test sets for adaptation.
}
\label{tab:efficiency}
\resizebox{\columnwidth}{!}{
\begin{tabular}{l||cc|cc|cc}
\toprule
\rowcolor[HTML]{E5E4E2}
\multicolumn{7}{c}{\textbf{Data efficiency analysis}} \\
\midrule
\multicolumn{1}{c||}{} & \multicolumn{2}{c|}{ColoredMNIST} & \multicolumn{2}{c|}{Office-Home} & \multicolumn{2}{c}{ImageNet-C} \\
\cmidrule{2-7}
\multicolumn{1}{c||}{\multirow{-2}{*}{Method}} & \%adapt & \%corr-adapt & \%adapt & \%corr-adapt & \%adapt & \%corr-adapt \\
\midrule
EATA & 15.6 & 10.6 & 29.7 & 14.5 & 11.7 & 7.0 \\
SAR & 19.6 & 12.1 & 31.8 & 18.9 & 13.1 & 8.2 \\
DeYO & 18.8 & 14.4 & 31.7 & 19.9 & 13.7 & 9.2 \\
\rowcolor[HTML]{ACE5EE}
\textbf{DualTTA} & \textbf{33.1} & \textbf{27.3} & \textbf{42.6} & \textbf{33.6} & \textbf{25.8} & \textbf{19.8} \\
\bottomrule
\end{tabular}
}
\end{table}
\subsubsection{Impacts of utilizing likely-incorrect samples.}
To evaluate the effectiveness of our proposed sample selection method, particularly the use of likely-incorrect samples, we compare our approach with a variant that only minimizes entropy on the likely-correct samples $\mathcal{D}^+$. The results in \Tref{tab:efficiency} show that the number of samples used for the adaptation process in the proposed method TTA is greater than that of all baseline methods DeYO and EATA. The performance of DualTTA is also improved 1.57\% when leveraging knowledge from likely-incorrect samples.

We further evaluate the impact of our proposed dual-optimization strategy by integrating it into existing baselines that utilize entropy-based sample selection, including EATA \cite{niu2022efficient} and DeYO \cite{lee2024entropy}. To achieve this, we define likely-correct and likely-incorrect sample sets based on the selection criteria established by these methods.
For DeYO, likely-correct samples are defined as:
\[
\mathcal{D}_{DeYO}^+ = \{ \x \mid \text{Ent}(\x) < \tau_{Ent},\ \text{PLPD}(\x,\x') > \tau_{\text{PLPD}} \},
\]
where $\x'$ represents a perturbed version of $\x$; the likely-incorrect samples are defined as:
\[
\mathcal{D}_{DeYO}^- = \{ \x \mid \text{Ent}(\x) < \tau_{Ent},\ \text{PLPD}(\x,\x') < \tau_{\text{PLPD}}/2 \},
\]
For EATA, the likely-correct sample set is given by:
\[
\mathcal{D}_{EATA}^+ = \{ \x \mid \text{Ent}(\x) < \tau_{Ent},\cos (\theta(\x),\mathbf{n}^{t-1})<\epsilon\},
\]
\[
\textrm{where }\hspace{3ex} \mathbf{n}^{t} =
\begin{cases} 
\bar{y}^{1}, & \text{if } t = 1 \\
\alpha \bar{y}^{t} + (1 - \alpha) \mathbf{n}^{t-1}, & \text{if } t > 1
\end{cases}
\]
Here, $\bar{y}^{t}$ represents the average model prediction for a batch of $B$ test samples at iteration $t$. The likely-incorrect samples are defined as:
\[
\mathcal{D}_{EATA}^- = \{ \x \mid \text{Ent}(\x) < \tau_{Ent},\cos (\theta(\x),\mathbf{n}^{t-1})>\frac{3}{2}\epsilon\}.
\]
\Tref{tab:impact_incorrect} shows that across all experimental scenarios, incorporating adaptation with likely-incorrect samples using our proposed method consistently enhances the performance of both DeYO and EATA compared to their original versions. The maximum performance improvement achieved by leveraging likely-incorrect samples reaches up to 1.19\% for DeYO and 3.48\% for EATA, highlighting the effectiveness of our approach.

\subsubsection{Impact of semantic altering, preserving threshold.}
We conduct experiments with different values of $\tau^{sa}$ and $\tau^{sp}$ across four datasets: ColoredMNIST, Waterbirds, PACS, and Office-Home, presented in \Fref{hyperparam}. We find that the model's performance remains relatively stable despite variations in $\tau^{sa}$ and $\tau^{sp}$. Notably, regardless of the threshold value, DualTTA consistently outperforms DeYO. To improve generalizability, we set $\tau^{sa}=0.4$, $\tau^{sp} = 0.7$ for all experimental scenarios. See the Supplementary for analysis on other hyper parameters.

\subsubsection{Efficient adaptation analysis.} \Tref{tab:efficiency} presents an analysis of adaptation efficiency across TTA methods, using two key metrics: \%adapt, which indicates the proportion of test samples selected for adaptation, and \%corr-adapt, which measures the accuracy of the selected samples. 
The visualization has been shown in \Fref{fig:sample_limitation2}.

Across all benchmarks, DualTTA substantially outperforms the baselines in both sample coverage and adaptation quality. On ImageNet-C, it adapts to 25.8\% of test samples, compared with 13.7\% of DeYO, while the share of correctly adapted samples rises to 19.8\%, versus 9.2\%. A similar gap is observed on Office-Home, where DualTTA improves coverage from 31.7\% to 42.6\% and correct-adapt from 19.9\% to 33.6\%. 

\section{Conclusions}
We introduced DualTTA, an innovative  framework designed to harness a larger and more diverse set of test samples for adaptation. Our approach pioneers a novel criterion that uses prediction stability to classify test samples as either likely-correct or likely-incorrect, based on their response to semantic-altering and semantic-preserving transformations.
We proposed a dual optimization strategy: reinforcing alignment for likely-correct samples through entropy minimization while applying corrective adjustments for likely-incorrect samples via entropy maximization. 
Extensive experiments across multiple datasets demonstrated that DualTTA surpasses state-of-the-art TTA methods, achieving a significant performance boost, with a gap of up to \textbf{8.06\%} over the second-best. 

\myheading{Acknowledgments.}{\small This project was initiated with support from the University of Adelaide's Global Engagement Fund. Subsequent support was provided in part by the Australian Institute for Machine Learning (University of Adelaide), the Centre for Augmented Reasoning (an initiative of the Department of Education, Australian Government), and Hanoi University of Science and Technology grant number T2024-TD-002.}

{
    \small
    \bibliographystyle{ieeenat_fullname}
    \bibliography{main}

@string{bmvc =  "Proceedings of the British Machine Vision Conference"}

@string{cvpr   = "Proceedings of the {IEEE} Conference on Computer Vision and Pattern Recognition"}

@string{iccv =  "Proceedings of the International Conference on Computer Vision"}

@string{icml =  "Proceedings of the International Conference on Machine Learning"}

@string{iclr ="Proceedings of International Conference on Learning and Representation"}

@string{ijcv =  "International Journal of Computer Vision"}

@string{nips = "Advances in Neural Information Processing Systems"}

@string{neurips = "Advances in Neural Information Processing Systems"}

@string{pami =  "IEEE Transactions on Pattern Analysis and Machine Intelligence"}

@string{wacv =  "Proceedings of the IEEE Workshop on Applications of Computer Vision"}

@inproceedings{hendrycks2019benchmark,
  author    = {Dan Hendrycks and Thomas Dietterich},
  title     = {Benchmarking Neural Network Robustness to Common Corruptions and Perturbations},
  booktitle = iclr,
  year      = {2019},
}

@inproceedings{koh2021wilds,
  author    = {Pang Wei Koh and Shiori Sagawa and Henrik Marklund and Sang Michael Xie and Marvin Zhang and Akshay Balsubramani and Weihua Hu and Michihiro Yasunaga and Richard L. Phillips and Irena Gao and others},
  title     = {WILDS: A Benchmark of In-the-Wild Distribution Shifts},
  booktitle = icml,
  year      = {2021},
}

@inproceedings{m_Zhang-Hoai-CVPR23,
author = {Zekun Zhang and Minh Hoai},
title = {Object Detection with Self-Supervised Scene Adaptation},
year = {2023},
booktitle = cvpr,
}

@inproceedings{m_Zhang-etal-BMVC24,
author = {Zekun Zhang and Vu Quang Truong and Minh Hoai},
title = {Efficiency-preserving Scene-adaptive Object Detection},
year = {2024},
booktitle = bmvc,
}

@inproceedings{liu2021ttt,
  author    = {Yuejiang Liu and Parth Kothari and Bastien van Delft and Baptiste Bellot-Gurlet and Taylor Mordan and Alexandre Alahi},
  title     = {TTT++: When Does Self-Supervised Test-Time Training Fail or Thrive?},
  booktitle = NeurIPS,
  year      = {2021}
}

@inproceedings{sun2020test,
  author    = {Yu Sun and Xiaolong Wang and Zhuang Liu and John Miller and Alexei Efros and Moritz Hardt},
  title     = {Test-Time Training with Self-Supervision for Generalization Under Distribution Shifts},
  booktitle = icml,
  year      = {2020},
}

@inproceedings{chen2022contrastive,
  author    = {Dian Chen and Dequan Wang and Trevor Darrell and Sayna Ebrahimi},
  title     = {Contrastive Test-Time Adaptation},
  booktitle = cvpr,
  year      = {2022}
}

@inproceedings{lee2024entropy,
  author    = {Jonghyun Lee and Dahuin Jung and Saehyung Lee and Junsung Park and Juhyeon Shin and Uiwon Hwang and Sungroh Yoon},
  title     = {Entropy is Not Enough for Test-Time Adaptation: From the Perspective of Disentangled Factors},
  booktitle = iclr,
  year      = {2024}
}

@inproceedings{niu2022efficient,
  author    = {Shuaicheng Niu and Jiaxiang Wu and Yifan Zhang and Yaofo Chen and Shijian Zheng and Peilin Zhao and Mingkui Tan},
  title     = {Efficient Test-Time Model Adaptation Without Forgetting},
  booktitle = icml,
  year      = {2022}
}

@inproceedings{niu2023towards,
  author    = {Shuaicheng Niu and Jiaxiang Wu and Yifan Zhang and Zhiquan Wen and Yaofo Chen and Peilin Zhao and Mingkui Tan},
  title     = {Towards Stable Test-Time Adaptation in Dynamic Wild World},
  booktitle = iclr,
  year      = {2023}
}

@inproceedings{wang2021tent,
  author    = {Dequan Wang and Evan Shelhamer and Shaoteng Liu and Bruno Olshausen and Trevor Darrell},
  title     = {Tent: Fully Test-Time Adaptation by Entropy Minimization},
  booktitle = iclr,
  year      = {2021}
}

@inproceedings{wang2022continual,
  author    = {Qin Wang and Olga Fink and Luc Van Gool and Dengxin Dai},
  title     = {Continual Test-Time Domain Adaptation},
  booktitle = cvpr,
  year      = {2022},
}

@inproceedings{cho2025feature,
  author    = {Younggeol Cho and Youngrae Kim and Junho Yoon and Seunghoon Hong and Dongman Lee},
  title     = {Feature Augmentation based Test-Time Adaptation},
  booktitle = wacv,
  year      = {2025}
}

@inproceedings{foret2020sharpness,
  author    = {Pierre Foret and Ariel Kleiner and Hossein Mobahi and Behnam Neyshabur},
  title     = {Sharpness-Aware Minimization for Efficiently Improving Generalization},
  booktitle = iclr,
  year      = {2021}
}

@inproceedings{lee2023towards,
  author    = {Jungsoo Lee and Debasmit Das and Jaegul Choo and Sungha Choi},
  title     = {Towards Open-Set Test-Time Adaptation Utilizing the Wisdom of Crowds in Entropy Minimization},
  booktitle = iccv,
  year      = {2023}
}

@inproceedings{lim2022ttn,
  author    = {Hyesu Lim and Byeonggeun Kim and Jaegul Choo and Sungha Choi},
  title     = {TTN: A Domain-Shift Aware Batch Normalization in Test-Time Adaptation},
  booktitle = iclr,
  year      = {2022}
}

@inproceedings{park2023label,
  author    = {Sunghyun Park and Seunghan Yang and Jaegul Choo and Sungrack Yun},
  title     = {Label Shift Adapter for Test-Time Adaptation Under Covariate and Label Shifts},
  booktitle = iccv,
  year      = {2023}
}

@inproceedings{wang2023feature,
  author    = {Shuai Wang and Daoan Zhang and Zipei Yan and Jianguo Zhang and Rui Li},
  title     = {Feature Alignment and Uniformity for Test Time Adaptation},
  booktitle = cvpr,
  year      = {2023}
}

@article{zhang2022memo,
  author    = {Marvin Zhang and Sergey Levine and Chelsea Finn},
  title     = {MEMO: Test Time Robustness via Adaptation and Augmentation},
  journal   = NeurIPS,
  volume    = {35},
  pages     = {38629--38642},
  year      = {2022}
}

@inproceedings{zhou2021domain,
  author    = {Kaiyang Zhou and Yongxin Yang and Yu Qiao and Tao Xiang},
  title     = {Domain Generalization with MixStyle},
  booktitle = iclr,
  year      = {2021}
}

@inproceedings{huang2017arbitrary,
  author    = {Xun Huang and Serge Belongie},
  title     = {Arbitrary Style Transfer in Real-Time with Adaptive Instance Normalization},
  booktitle = iccv,
  year      = {2017}
  }

@inproceedings{yun2019cutmix,
  title={CutMix: Regularization strategy to train strong classifiers with localizable features},
  author={Yun, Sangdoo and Han, Dongyoon and Oh, Seong Joon and Chun, Sanghyuk and Choe, Junsuk and Yoo, Youngjoon},
  booktitle=iccv,
  year={2019}
}

@inproceedings{cubuk2020randaugment,
  title={RandAugment: Practical automated data augmentation with a reduced search space},
  author={Cubuk, Ekin D and Zoph, Barret and Shlens, Jonathon and Le, Quoc V},
  booktitle=NeurIPS,
  year={2020}
}

@inproceedings{yang2020fda,
  title={FDA: Fourier domain adaptation for semantic segmentation},
  author={Yang, Yanchao and Soatto, Stefano},
  booktitle=cvpr,
  year={2020}
}

@inproceedings{gatys2016image,
  title={Image style transfer using convolutional neural networks},
  author={Gatys, Leon A and Ecker, Alexander S and Bethge, Matthias},
  booktitle=cvpr,
  year={2016}
}

@inproceedings{jackson2019style,
  title={Style Augmentation: Data Augmentation via Style Randomization},
  author={Jackson, Philip T and Atapour-Abarghouei, Amir and Bonner, Stephen and Breckon, Toby P and Obara, Boguslaw},
  booktitle={Proceedings of the IEEE/CVF Conference on Computer Vision and Pattern Recognition Workshops},
  year={2019}
}

@inroceedings{pmlr-v119-sun20b,
  title = 	 {Test-Time Training with Self-Supervision for Generalization under Distribution Shifts},
  author =       {Sun, Yu and Wang, Xiaolong and Liu, Zhuang and Miller, John and Efros, Alexei and Hardt, Moritz},
  booktitle = 	 {Proceedings of the 37th International Conference on Machine Learning},
  year = 	 {2020}
}

@article{shannon2001mathematical,
  author    = {Claude Elwood Shannon},
  title     = {A Mathematical Theory of Communication},
  journal   = {The Bell System Technical Journal},
  year      = {1948}
}

@article{zhou2022domain,
  title={Domain Generalization: A Survey},
  author={Zhou, Kaiyang and Liu, Ziwei and Qiao, Yu and Xiang, Tao and Loy, Chen Change},
  journal=pami,
  year={2022},
  publisher={IEEE},
}

@article{farahani2020brief,
  title={A Brief Review of Domain Adaptation},
  author={Farahani, Abolfazl and Voghoei, Sahar and Rasheed, Khaled and Arabnia, Hamid R.},
  journal={arXiv preprint arXiv:2010.03978},
  year={2020},
}

@article{sun2015survey,
  title={{A survey of multi-source domain adaptation}},
  author={Sun, Shiliang and Shi, Honglei and Wu, Yuanbin},
  journal={Information Fusion},
  volume={24},
  pages={84--92},
  year={2015},
  publisher={Elsevier}
}

@article{liang2024comprehensive,
  title={{A Comprehensive Survey on Test-Time Adaptation under Distribution Shifts}},
  author={Liang, Jian and He, Ran and Tan, Tieniu},
  journal=ijcv,
  volume={133},
  number={1},
  pages={31--64},
  year={2024},
  publisher={Springer}
}

@inproceedings{li2017deeper,
  title={{Deeper, Broader and Artier Domain Generalization}},
  author={Li, Da and Yang, Yongxin and Song, Yi-Zhe and Hospedales, Timothy M.},
  booktitle=iccv,
  year={2017}
}

@inproceedings{venkateswara2017deep,
  title={{Deep Hashing Network for Unsupervised Domain Adaptation}},
  author={Venkateswara, Hemanth and Eusebio, Jose and Chakraborty, Shayok and Panchanathan, Sethuraman},
  booktitle=cvpr,
  year={2017}
}

@inproceedings{li2021on,
  author    = {Li, Boyi and Wu, Felix and Lim, Ser-Nam and Belongie, Serge and Weinberger, Kilian Q.},
  title     = {On Feature Normalization and Data Augmentation},
  booktitle = cvpr,
  year      = {2021}
}

@inproceedings{wang2019transferable,
  title={Transferable Normalization: Towards Improving Transferability of Deep Neural Networks},
  author={Wang, Ximei and Jin, Ying and Long, Mingsheng and Wang, Jianmin and Jordan, Michael I.},
  booktitle=nips,
  year={2019}
}

@inproceedings{gao2021representative,
  title={Representative Batch Normalization with Feature Calibration},
  author={Gao, Shang-Hua and Han, Qi and Li, Duo and Cheng, Ming-Ming and Peng, Pai},
  booktitle=cvpr,
  year={2021}
}

@inproceedings{tran2025conststyle,
  title={ConstStyle: Robust Domain Generalization with Unified Style Transformation},
  author={Tran, Nam Duong and Phuong, Nam Nguyen and Pham, Hieu H and Le Nguyen, Phi and Thai, My T},
  booktitle={Proceedings of the IEEE/CVF International Conference on Computer Vision},
  pages={3174--3183},
  year={2025}
}
}


\end{document}


\onecolumn
\section{Verification for Style-altering criteria}
\noindent \textbf{Preliminaries.}
Let $x$ be an input sample, and let $z = \phi(x) \in \mathbb{R}^{C \times H \times W}$ denote the feature maps before batch normalization (BN), computed channel-wise.

For BN, the channel-wise statistics are denoted as
\[
\mu = (\mu_c)_{c=1}^C, \quad \sigma = (\sigma_c)_{c=1}^C,
\]
computed over spatial dimensions (e.g., per-sample or per-batch mean/std).

A \textit{style-altering} transformation perturbs the BN statistics as
\[
\mu^{sa} = \mu + \Delta\mu, \quad \sigma^{sa} = \sigma + \Delta\sigma.
\]
The transformed feature maps are obtained channel-wise by:
\[
z_c^{sa} = \frac{z_c - \mu_c}{\sigma_c} \sigma_c^{sa} + \mu_c^{sa}.
\]

Let $s(x; \theta, \mu, \sigma) \in \mathbb{R}^K$ denote the classifier logits (where $s_k$ is the logit for class $k$).  
Probabilities are given by $p_k = \mathrm{softmax}(s)_k$.  
Here, $\theta$ represents all model parameters except the affine BN parameters.

The predicted label is:
\[
\hat{y}(x) = \arg\max_k s_k(x).
\]

We define the \textbf{margin} of a sample as:
\[
m(x) = s_y(x) - \max_{k \neq y} s_k(x), 
\quad \text{where } y = \hat{y}(x).
\]

\noindent \textbf{Theorem 1.}
Given a sample $x$, assume there exists a small perturbation $(\Delta\mu, \Delta\sigma)$ that makes the label of $x$ changes when turn into $x^{sa}$, i.e.
$
\hat{y}(x^{sa}) \neq \hat{y}(x)
$. 
Then, exist $K$ such that:
\[
m(x) \le 
K \cdot (\|\Delta\mu\|
+ \|\Delta\sigma\|)
+ o(\|\Delta\|).
\]

\noindent \textbf{Proof of theorem 1.}
The label of $x$ changes when turn into $x^{sa}$, i.e.
$
\hat{y}(x^{sa}) \neq \hat{y}(x)
$. 
meaning there exists a class $k$ satisfying:
\[
s_k(x^{sa}) > s_y(x^{sa}).
\]

Using the first-order Taylor expansion of logits with respect to the BN parameters (assuming differentiability of $s_k$):
\[
s_k(x^{sa}) \approx s_k(x) 
+ \nabla_{\mu}s_k(x) \cdot \Delta\mu 
+ \nabla_{\sigma}s_k(x) \cdot \Delta\sigma 
+ o(\|\Delta\|).
\]

The \textbf{margin} $m(x) = s_y(x) - s_k(x)$.  
A label flip occurs when:
\[
m(x) + (\nabla_{\mu}s_y - \nabla_{\mu}s_k) \cdot \Delta\mu
+ (\nabla_{\sigma}s_y - \nabla_{\sigma}s_k) \cdot \Delta\sigma < 0.
\]

Hence, if the flip happens under a small perturbation, it must hold that:
\[
m(x) \le |
(\nabla_{\mu}s_y - \nabla_{\mu}s_k) \cdot \Delta\mu
+ (\nabla_{\sigma}s_y - \nabla_{\sigma}s_k) \cdot \Delta\sigma
+ o(\|\Delta\|)|.
\]
Assume that $s_y-s_k$ is a $K-$Lipschitz function then there exist $K$ such that $\max(\nabla_\mu(s_y-s_k),\nabla_\sigma(s_y-s_k))=K$, then 
\[
|
(\nabla_{\mu}s_y - \nabla_{\mu}s_k) \cdot \Delta\mu
+ (\nabla_{\sigma}s_y - \nabla_{\sigma}s_k) \cdot \Delta\sigma
+ o(\|\Delta\|)|\le K\cdot(\|\Delta\mu||+\|\Delta\sigma||)+o(\|\Delta\|)|.
\]
Hence, 
\[
m(x) \le K\cdot(\|\Delta\mu||+\|\Delta\sigma||)+o(\|\Delta\|)|.
\]
\noindent
\textbf{In summary:}  
The initial margin $m(x)$ must be very small relative to the gradient of the logits with respect to BN statistics in the direction of the perturbation.  
This implies that the sample lies close to the decision boundary along the style-altering direction.
\section{Rationality of Combining the Two Criteria}
Let
\begin{itemize}
  \item $h=h(x;\theta)$ be an intermediate representation (feature vector) produced by the model, and assume (locally) that it decomposes as
  \[
    h = \begin{bmatrix} h_c \\ h_s \end{bmatrix},
    \qquad h_c\in\mathbb{R}^{d_c},\quad h_s\in\mathbb{R}^{d_s},
  \]
  where $h_c$ is the \emph{content} subvector and $h_s$ is the \emph{style} subvector.
  \item $p(x)=p(\cdot\mid x;\theta)\in\mathbb{R}^C$ be the network softmax output (column vector).
  \item $J_{h_s,\theta}=\dfrac{\partial h_s}{\partial\theta}\in\mathbb{R}^{d_s\times m}$ be the Jacobian of $h_s$ w.r.t.\ parameters $\theta\in\mathbb{R}^m$.
  \item $J_s(x)=\dfrac{\partial p}{\partial h_s}\in\mathbb{R}^{C\times d_s}$ denote the \emph{style-Jacobian}, i.e. the Jacobian of the outputs $p$ with respect to the style representation $h_s$.
\end{itemize}


By the chain rule,
\[
\nabla_\theta L(x) \;=\; \frac{\partial L}{\partial p}\,\frac{\partial p}{\partial \theta}
\;=\; \bigg(\frac{\partial p}{\partial \theta}\bigg)^\top \frac{\partial L}{\partial p},
\]
where $\dfrac{\partial p}{\partial \theta}\in\mathbb{R}^{C\times m}$ is the Jacobian of $p$ w.r.t.\ $\theta$.
Since $p$ depends on $\theta$ through $h$, we further expand
\[
\frac{\partial p}{\partial \theta} \;=\; \frac{\partial p}{\partial h}\,\frac{\partial h}{\partial \theta}
\;=\; J_{h,p}\,J_{h,\theta},
\]
and equivalently in the transposed form used below,
\[
\nabla_\theta L \;=\; J_{h,\theta}^\top \,\nabla_h L,
\qquad\text{with}\quad
\nabla_h L \;=\; \bigg(\frac{\partial L}{\partial p}\bigg)^\top \!J_{h,p}.
\]

Partitioning by content/style,
\[
J_{h,\theta} \;=\; \begin{bmatrix} J_{h_c,\theta} \\[4pt] J_{h_s,\theta} \end{bmatrix},\qquad
\nabla_h L \;=\; \begin{bmatrix} \nabla_{h_c} L \\[4pt] \nabla_{h_s} L \end{bmatrix},
\]
we obtain the decomposition
\[
\nabla_\theta L \;=\; J_{h_c,\theta}^\top \nabla_{h_c} L \;+\; J_{h_s,\theta}^\top \nabla_{h_s} L.
\]

\section{Definition of the style-gradient component}
Define the \emph{style component} of the parameter gradient by
\[
\boxed{ \; g_s \;:=\; J_{h_s,\theta}^\top \,\nabla_{h_s} L \; \in \mathbb{R}^m \; }.
\]
Since $L$ depends on $p$, we apply the chain rule to $\nabla_{h_s} L$:
\[
\nabla_{h_s} L \;=\; \Big(\frac{\partial L}{\partial p}\Big)^\top \! J_s(x),
\]
where $J_s(x)=\partial p/\partial h_s$ (size $C\times d_s$). Combining yields the commonly used identity
\[
\boxed{ \; g_s \;=\; J_{h_s,\theta}^\top \; \Big( J_s(x)^\top \; \frac{\partial L}{\partial p} \Big) \;=\; J_{h_s,\theta}^\top \, \nabla_{h_s} L. \; }
\]

This equation expresses how sensitivity of outputs to style ($J_s$) and sensitivity of style features to parameters ($J_{h_s,\theta}$) jointly produce the parameter-space style-gradient $g_s$.




For a small style perturbation $\Delta h_s$,
\[
\Delta p \approx J_s(x)\,\Delta h_s.
\]
The resulting first-order loss change is
\[
\Delta_{h_s} L \approx \Big(\frac{\partial L}{\partial p}\Big)^\top J_s(x)\,\Delta h_s.
\]
Thus, if $\|J_s(x)\|$ is large, even small $\Delta h_s$ induce large $\Delta p$ and hence large $\Delta L$; this corresponds to a large gradient $\nabla_{h_s}L$ and, after multiplication by $J_{h_s,\theta}^\top$, to a large $g_s$.